\pdfoutput=1
\documentclass[11pt]{article}

\usepackage[final]{acl}

\usepackage{times}
\usepackage{latexsym}

\usepackage[T1]{fontenc}
\usepackage[utf8]{inputenc}

\usepackage{microtype}

\usepackage{inconsolata}

\usepackage{graphicx}
\usepackage{multirow}
\usepackage{xcolor}
\title{Enhanced Detection of Conversational Mental Manipulation Through Advanced Prompting Techniques}

\author{
Ivory Yang$^1$ \quad 
Xiaobo Guo$^2$ \quad 
Sean Xie$^3$ \quad 
Soroush Vosoughi$^4$ \\
$^{1,2,3,4}${Department of Computer Science, Dartmouth College} \\
$^1${\texttt{Ivory.Yang.GR@dartmouth.edu}}\\
$^4${\texttt{Soroush.Vosoughi@dartmouth.edu}}
}

\begin{document}
\maketitle
\begin{abstract}
This study presents a comprehensive, long-term project to explore the effectiveness of various prompting techniques in detecting dialogical mental manipulation. We implement Chain-of-Thought prompting with Zero-Shot and Few-Shot settings on a binary mental manipulation detection task, building upon existing work conducted with Zero-Shot and Few-Shot prompting. Our primary objective is to decipher why certain prompting techniques display superior performance, so as to craft a novel framework tailored for detection of mental manipulation. Preliminary findings suggest that advanced prompting techniques may not be suitable for more complex models, if they are not trained through example-based learning.
\end{abstract}

\section{Introduction}
Mental manipulation is a subtle form of psychological influence on preferences and choice \cite{barnhillManipulation2014, bublitz2014crimes}, and its impact on society has been exacerbated by advancements in technology \cite{carroll2023characterizing}. The detection of such manipulative language poses a significant hurdle within the field of Natural Language Processing (NLP) \cite{huffaker2020crowdsourced}, due to its subtle, context-dependent and inherently nuanced nature. To address these challenges, we investigate the effectiveness of advanced prompting techniques at detecting mental manipulation. The results of our xexperiments show that while Chain-of-Thought (CoT) prompting achieves exceptional accuracy across scenarios, special attention must be given to the learning configuration in order to ensure its high performance. 

\begin{figure}[t]
  \centering
  \includegraphics[width=\linewidth]{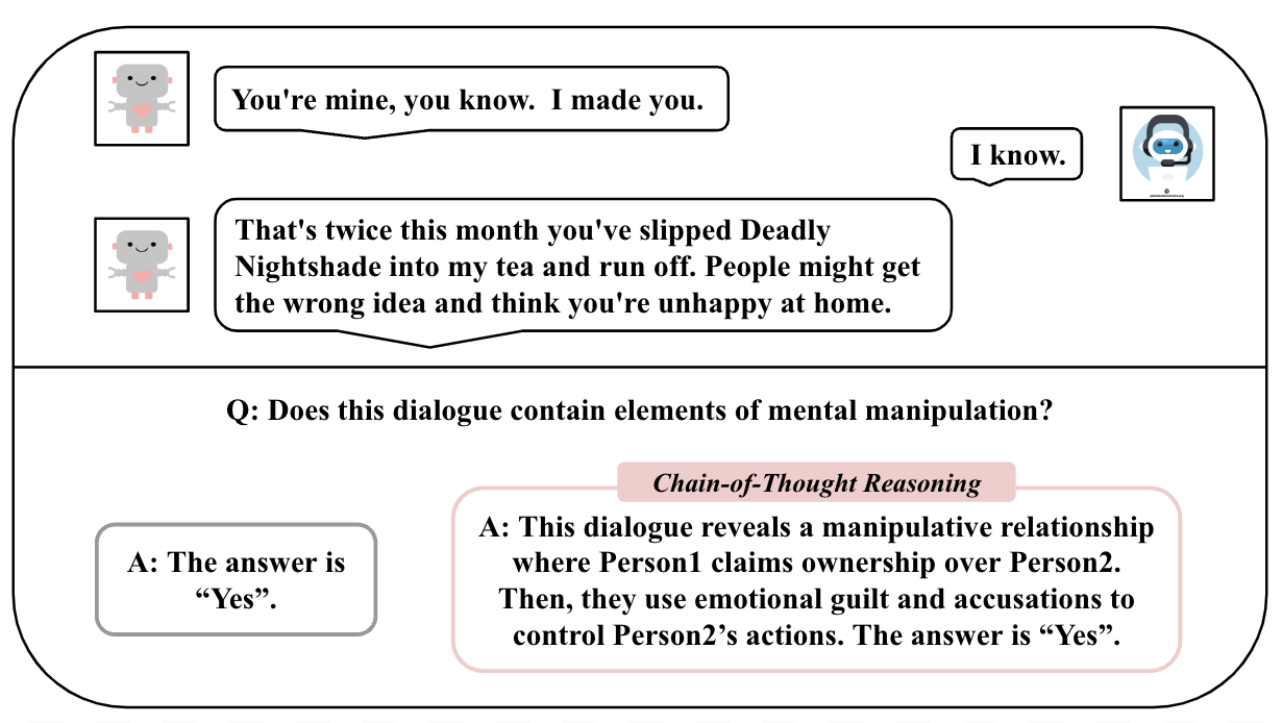}
  \caption{An example dialogue incorporating Chain-of-Thought reasoning in mental manipulation detection}
  \label{fig:illustration}
\end{figure}

\section{Related Work}
~\citet{wang-etal-2024-mentalmanip} proposed MentalManip, a dataset comprising of over 4000 human-annotated dialogues focused on mental manipulation. By leveraging Zero-Shot and Few-Shot prompting, the authors set a foundational understanding of model limitations in scenarios devoid of explicit toxic indicators. In recent years, there has been an advent in sophisticated prompting techniques to improve LLM performance on complex NLP tasks. Chain-of-Thought prompting \cite{wei2022chain} uses example-based learning for models to generate intermediate reasoning steps before arriving at a conclusion. As an extension, Zero-Shot Chain-of-Thought \cite{kojima2022large} relies solely on crafted prompts that guide the model to articulate its gradiational reasoning process. This paper aims to apply these advanced prompting techniques to the problem of mental manipulation detection.

{
\begin{table*}[!ht]
    \centering
    \small
    \resizebox{\textwidth}{!}{%
    \begin{tabular}{c|ccccc|ccccc}
    \hline
    Experiment Setting & \multicolumn{5}{c|}{GPT-3.5} & \multicolumn{5}{c}{GPT-4o}
    % \\ \cmidrule(lr){3-7}\cmidrule(lr){8-12}
    \\ 
    & $P$ & $R$ & $Acc$ & $F_1^{mi}$ & $F_1^{ma}$ & $P$ & $R$ & $Acc$ & $F_1^{mi}$ & $F_1^{ma}$
    \\ \hline
    Zero-Shot & $.750$ & $.827$ & $.693$ & $.693$ & $.620$ & $.741$ & $.952$ & $.739$ & $.739$ & $.617$
    \\ \hline
    Few-Shot & $.789$ & $.769$  & $.702$ & $.702$ & $.659$ & $.749$  & $.980$  & $.762$  & $.762$ & $.641$
    \\ \hline
    CoT (Zero) & $.725$ & $.951$  & $.721$ & $.721$ & $.579$ & $.729$  & $.950$  & $.724$  & $.724$ & $.586$
    \\ \hline
    CoT (Few) & $.714$ & $.909$  & \boldmath{$.722$} & $.722$ & $.673$ & $.769$  & $.909$  & \boldmath{$.778$}  & $.778$ & $.750$
                    
    \\ \hline

    \end{tabular}
    }
    \vspace{-2mm}
    \caption{Results of manipulation detection task on MentalManipCon dataset. $P$, $R$, $Acc$, $F_1^{mi}$, and $F_1^{ma}$ stands for binary precision, binary recall, accuracy, micro $F_1$, and macro $F_1$ respectively. ``CoT (Zero)'': ``Chain-of-Thought with Zero-Shot learning settings'', ``CoT (Few)'': ``Chain-of-Thought with Few-Shot learning settings''}
    \label{tab:exp_1}
\end{table*}
}

\section{Experiment}
\subsection{Setup}
We conducted our experiments using the \textbf{MentalManipCon} dataset \cite{wang-etal-2024-mentalmanip}, a rigorously selected corpus of dialogues on mental manipulation, with consensus agreement amongst all three reviewers. We then conducted our experiments across two LLMs: GPT-3.5 and GPT-4o, to assess their performance in detecting mental manipulation. Both models were evaluated using the same set of tasks, with effectiveness measured by Precision, Recall, Accuracy and F1-Score across prompting strategies. Due to binary classification, Accuracy and Micro F1 reflect the same scores.

\subsection{Prompting Techniques}
In our study, we assess the effectiveness of four different prompting strategies in identifying mental manipulation: Zero-Shot, Few-Shot and CoT with Zero-Shot and Few-Shot settings. The Zero-Shot approach was implemented without providing any prior examples to the model. In contrast, the Few-Shot approach utilized a set of three examples (two manipulative, one non-manipulative) which were randomly chosen to guide the model. The CoT strategy with Zero-Shot settings involved a modified prompt that encouraged the model to process its thoughts step-by-step, integrating a reasoning component into its responses. For CoT with Few-Shot settings, we enlisted two college students, both native English speakers, to manually annotate detailed, step-by-step reasoning for a randomly selected set of 42 examples. This annotated dataset was then employed to train our model for CoT prompting.

\section{Results}
From the results in Table~\ref{tab:exp_1}, it is evident that Few-Shot CoT produces the best performance in terms of Accuracy, at 0.722 for GPT-3.5 and 0.778 for GPT-4o, which aligns with our expectations. However, it is notable that when upgrading from GPT-3.5 to GPT-4o, Zero-Shot CoT drops from second best performing to the worst performing technique, even worse than regular Zero-Shot. It also produces a significant number of false positives, as reflected by its Macro F1 scores, which are the lowest across all techniques for both GPT-3.5 and GPT-4o. 

% The exceptional performance of Few-Shot CoT can be attributed to its structured reasoning process, closely mirroring human cognitive strategies. However, results suggest that {\em although CoT shows better performance in general, on the task of mental manipulation, it should be combined with the samples}. In a Zero-Shot setting without specific task training, the increased complexity of the model might cause over-generalization, and its advanced reasoning may construct more convincing rationales for incorrect classifications. A flawed step-by-step reasoning could also reflect biased pre-training.

The exceptional performance of Few-Shot CoT can be attributed to its structured reasoning process, closely mirroring human cognitive strategies. However, by comparing the performance of Zero-Shot settings, we observe that although CoT shows better performance in general, on the task of mental manipulation, it should be combined with samples for learning. Considering the reduced Precision on both GPT-3.5 and GPT-4o with Zero-Shot CoT, we assume that both models may wrongly understand the definition of mental manipulation, which is further enhanced during CoT. Therefore, we conduct a pilot check on the generated reasons.

From manually checking results, GPT-4o places overemphasis on verbal cues and misinterprets fragmented or informal speech. The model attributes manipulation to communication style rather than actual manipulative intent. GPT-4o also appears to be biased towards conflict, detecting manipulation even in benign situations, and interpreting neutral or vague responses as signs of manipulation. For mental manipulation detection, CoT \textbf{\underline{without example-based learning}} may \textbf{\underline{perform worse}} in relation to simpler techniques as \textbf{\underline{model complexity increases}}.

\section{Future Work}
We aim to extend our analysis of CoT performance concerning model complexity, and the definition of mental manipulation provided to the model. Moreover, we will explore other prompting techniques such as Iterative prompting \cite{wang2022iteratively}, Self-Consistency \cite{wang2022self}, and Tree-of-Thoughts prompting \cite{long2023large, yao2024tree}. We also seek to take into account serial position effects \cite{guo2024serial} to analyze how placement of information within prompts affects detection accuracy. Given that this is a long-term project on mental manipulation detection, we may potentially expand our scope to consider the impact of gender \cite{grieve2019masculinity}, stereotypes \cite{ma2023deciphering} and biases \cite{xie2024addressing} in our results.

\bibliography{custom}

\end{document}